%% file: main.tex
\documentclass{IOS-Book-Article}

\usepackage{soul}\setuldepth{article}
\usepackage{times}
\normalfont
\usepackage{algorithm}
\usepackage[noend]{algpseudocode}
\usepackage{standalone}
\usepackage{import}
\usepackage{calc}
\usepackage{tikz}
\usepackage{subcaption}
\usepackage{wrapfig}

\import{figures/}{tikzcommon}

\def\hb{\hbox to 11.5 cm{}}

\newcommand{\add}[1]{#1}

\usepackage{researchpack}
\usepackage{visualsemanticpack}
\usepackage{xspace}
\usepackage{booktabs}
\usepackage[capitalise,nameinlink]{cleveref}

\newcommand{\hierarchy}{\ensuremath{\calH}\xspace}

\renewcommand*\Call[2]{\textproc{#1}(#2)} 

\begin{document}

\pagestyle{headings}
\def\thepage{}
\begin{frontmatter}              

\title{Egocentric Hierarchical Visual Semantics}


\author[A]{\fnms{Luca} \snm{Erculiani}},
\author[A]{\fnms{Andrea} \snm{Bontempelli}\thanks{Corresponding Author: Andrea Bontempelli, andrea.bontempelli@unitn.it}},
\author[A]{\fnms{Andrea} \snm{Passerini}},
and
\author[A]{\fnms{Fausto} \snm{Giunchiglia}}

\runningauthor{L. Erculiani et al.}
\address[A]{University of Trento}


\vspace{-0.1cm}
\begin{abstract}
We are interested in aligning how people think about objects and what machines perceive, meaning by this the fact that object recognition, as performed by a machine, should follow a process which resembles that followed by humans when thinking of an object associated with a certain concept. The ultimate goal is to build systems which can meaningfully interact with their users, describing  what they perceive \textit{in the users' own terms}.
As from the field of \textit{Lexical Semantics}, humans organize the meaning of words in hierarchies where the meaning
of, e.g., a noun, is defined in terms of the meaning of a more general noun, its \textit{genus}, and of one or more differentiating properties, its \textit{differentia}. 
The main tenet of this paper is that object recognition should implement a hierarchical process which follows the hierarchical semantic structure used to define the meaning of words. 
We achieve this goal by implementing an algorithm which, for any object, recursively recognizes its \textit{visual genus} and its \textit{visual differentia}. In other words, the recognition of an object is decomposed in a sequence of steps where the locally relevant visual features are recognized. This paper presents the algorithm and a first evaluation.
\end{abstract}

\vspace{-0.1cm}
\end{frontmatter}

\vspace{-0.3cm}
\section{Introduction}

\textit{Lexical Semantics} studies how word meanings, i.e., linguistic concepts~\cite{kuang2018integrating,UKC-IJCAI} are formed, where these concepts are assumed to be constructed by humans through language. As from this field, humans organize the meaning of words in hierarchies where the meaning
of, e.g., a noun,  is defined in terms of a more general noun, its \genus, and of one or more differentiating properties, its \differentia. Thus for instance, a \textit{guitar} is a \textit{stringed (musical) instrument} with \textit{six strings}~\cite{miller1990introduction}. 
The main tenet of the work described in this paper is that object recognition should implement a process which progressively visually reconstructs the hierarchical semantic structure used to define the meaning of words. Only in this way it is possible to have a full one-to-one \emph{alignment} between how people think of the world and, ultimately, human language, and machine perception.
The ultimate goal is to build systems which can meaningfully interact with their users, describing  what they perceive \textit{in the users' own terms}. Notice how this is a well known, still unsolved problem,  the so called  \textit{Semantic Gap problem}, which was identified in 2010
\cite{2010smeulders} as (quote) ``\textit{... the lack of coincidence between the information that one can extract from the visual data and the interpretation that the same data have for a user in a given situation.}".

Based on the work in the field of \textit{Teleosemantics}  \cite{macdonald2006teleosemantics}, see in particular the work in \cite{millikan1984language,millikan1998more,millikan2000clear,millikan2005language}, the field of \textit{Visual
Semantics} has been introduced as the study of how humans build concepts when using vision to perceive objects in the world~\cite{giunchiglia2021towards}.
According to this line of work, objects should be recognized by recognizing first their genus and then their differentia, as visually represented in the input images or videos. Thus, for instance, a \textit{guitar} should be recognized first as a \textit{stringed instrument} (\genus), which is itself a \textit{musical instrument} with \textit{strings}, and then by recognizing its \textit{six strings} (\differentia)~\cite{miller1990introduction,UKC-IJCAI}. This clearly leads to a recursive recognition process where the set of possible objects gets progressively restricted to satisfy more and more refined differentiae. 
In the most general case, the root node is the concept \textit{object} itself, namely anything that can be detected as such, e.g., via a bounding box. \add{In this context, we adopt an egocentric point-of-view with respect to a specific person \cite{erculiani2019continual,smithEgocentric}.}

\begin{figure}[t]
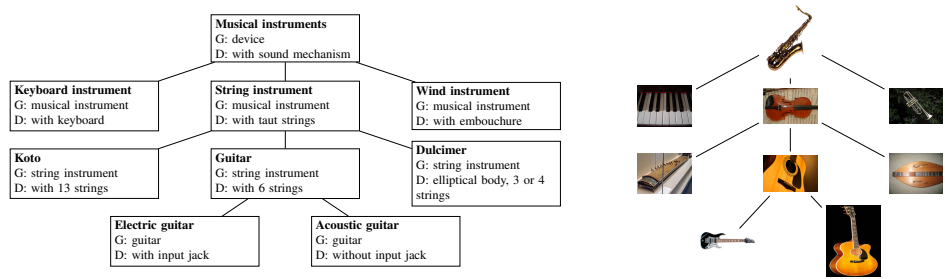

    \centering
    \includestandalone[width=1\textwidth]{figures/music/music}
    \vspace{-0.1cm}
    \caption{A classification concept hierarchy for Musical Instruments~\cite{2023-iConf-Mayuch}. \textbf{Left}: \textit{Lexical semantic} hierarchy and \genus (G) and \differentia (D) of each concept. \textbf{Right}: \textit{Visual semantic} hierarchy.} 
    \label{fig:lexsem}    
    \vspace{-0.2cm}
\end{figure}

As an example consider~\cref{fig:lexsem}, taken from a small classification of musical instruments~\cite{2023-iConf-Mayuch}. In this figure (left), we can see how the meaning of each label is provided in terms of a genus and a differentia, and where the genus one level down is the label of the concept the level up. Dually in the figure (right), we can see how all the images clearly show the differentia that allows to differentiate the object one level down from the object one level up (as having an extra feature, i.e., the visual differentia) and from all the siblings (as all objects under the same visual genus have a different visual differentia).
Notice how, in current hierarchical computer vision tasks, the hierarchy is usually a-priori and static, e.g., \cite{bertinetto2020making}, and does not consider the users' language and its mapping to their visual perception (see, e.g.,~\cite{lee2018hierarchical,ruiz2022hierarchical}), leading therefore to a human-machine misalignment.
For instance, a non-expert user would correctly classify the Koto instrument in~\cref{fig:lexsem} as a stringed instrument but, differently from a musician, would not describe it using its name, thus having two different but consistent linguistic descriptions of the same image.

The main goal of this paper presents a general algorithm which aligns machine perception and human description. 
The algorithm is based on two key ideas:
\begin{itemize}
    \item Object recognition is implemented following a \textit{hierarchical decomposition} process where the uniquely identifying features of the input object (the differentia) are recognized following the same order that is used in constructing the meaning of the label naming the object. 
    \item Object recognition follows an \textit{egocentric, incremental approach} where the user progressively refines the level of detail at which an object is recognized.
\end{itemize}
\noindent
The work in~\cite{giunchiglia2021towards} introduced visual semantics in the base hierarchy-less case. This paper extends this work to hierarchies of any depth. We do this by leveraging \textit{Extreme Value Machines}~\cite{rudd2018extreme}, a principled approach to open set problems which allows to implement differentia-based object recognition.
The source code of the algorithm, the dataset and all the material necessary to reproduce the experiments are freely available online.\footnote{\url{https://github.com/lucaerculiani/hierarchical-objects-learning}.}

\vspace{0.2cm}
\section{Visual Semantics}
\label{S2}

\begin{figure}[tb]
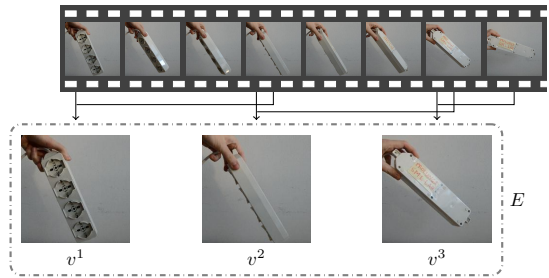

    \centering
     \includestandalone[width=0.6\textwidth]{figures/hyp2}
     \vspace{-0.1cm}
     \caption{\label{fig:encounter}Example of an encounter. The video
       contains eight frames of a power strip
       gradually rotated over time on a white background. Similar
       adjacent frames are aggregated in three visual objects, which
       form the encounter \enc~\cite{giunchiglia2021towards}. For
       better visualization, each visual object is represented, here
       and below, as its first frame.}
       \vspace{-0.2cm}
\end{figure}

\begin{figure}[tb]
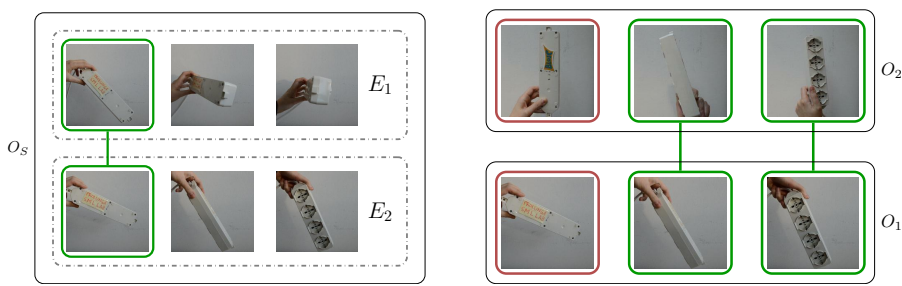

     \centering
     \includestandalone[width=0.45\linewidth]{figures/hyp1}
     \hspace{0.05\textwidth}
     \includestandalone[width=.45\linewidth]{figures/hyp3} 
     \vspace{-0.1cm}
     \caption{\label{fig:object}\textbf{Left}: An object made of two distinct encounters (dashed line), with their similar visual objects connected in green. \textbf{Right}: Two distinct objects sharing the same genus (via the visual objects connected in green). Red visual objects are the differentia (i.e., different tape on the back side).}
  \vspace{-0.2cm}
\end{figure}

We inherit from~\cite{giunchiglia2021towards} the following foundational notions.
An {\em encounter} \enc is an event during which a user sees an object. We computationally model an encounter as one or
more {\em visual objects}, where a visual object consists of a \textit{sequence of adjacent frames} that are
similar to each other. 
\cref{fig:encounter} shows an example of an encounter
with its decomposition into visual objects.
An {\em object} \obj is a collection of encounters that are perceived to
represent the same concept. 

The left part of
\cref{fig:object} shows an object consisting of two encounters, with the visual
objects that make the two encounters similar highlighted in green.
Two encounters that have at least a pair of visual objects that are
similar are said to share the same \genus. Two encounters with the same \genus could
or could not be associated with the same object. What makes this decision is the
presence (or the absence) of a \differentia, {i.e.}, a pair of visual
objects that identifies the two encounters as representing two
distinct objects. The right part of \cref{fig:object} visually
presents these concepts. The visual objects connected in green
indicate that the two objects share the same \genus, while those
circled in red are their \differentia, and indicate that they are
distinct objects.  The intuition is that some partial views of the
objects determine their \genus and \differentia respectively. The
hierarchy \hierarchy organizes the objects, modeled as visual objects,
in a tree-like structure, which outlines the subsumption relationships
between the objects in terms of \genus and \differentia (see, e.g.,
\cref{fig:lexsem}).

\section{Building an Egocentric Visual Semantic Hierarchy}
\label{sec:alg}

The visual hierarchy is built incrementally as the
objects are perceived. The interaction with the user ensures that the
visual hierarchy matches the user lexical semantics. The proposed
framework consists of a cyclic procedure in which at each iteration
a new encounter (a sequence depicting an object) is shown to the
model. The model then asks the user a series of queries over the
\genus and \differentia of the new encounter with respect to some of
the objects that were seen in the past by the algorithm (which are
stored in its internal memory). Via this interaction, the user can
guide the algorithm to assign the new encounter to the correct
position inside the machine's knowledge base.

\begin{algorithm}[t]
	\raggedright
   \begin{algorithmic}[1]
      \Procedure{main}{}
      

          \While {True}
            \State $\enc \gets \textsc{perceive}()$
            \State $\obj_e \gets \Call{predictGenus}{\enc}$
            \While{$not \; \Call{genusOf}{\enc, \obj_e}$}
                \State $\obj_e \gets \Call{parent}{\obj_e}$
            \EndWhile
            \State $\Call{refineGenus}{\obj_e, \enc}$
        \EndWhile
      \EndProcedure
    \end{algorithmic}
    \caption{The main loop of the framework.}
    \label{algo:mainloop}
\end{algorithm}

\paragraph{The main loop.} \cref{algo:mainloop} lists the
pseudo-code of the infinite learning loop that takes as input a new
sequence at each iteration. This new sequence is first forwarded to an
embedding algorithm that converts the video sequence, currently
encoded as a series of frames, into a collection of visual objects
(i.e. the encounter \enc). In this step, we employ an unsupervised deep neural network, pre-trained on a self-supervised class-agnostic task~\cite{caron2019unsupervised}. \add{This training approach ensures that the embeddings are not explicitly biased towards the classes.}
Then, the \textsc{PredictGenus} procedure searches
in its memory for the most specific \genus $\obj_e$ for encounter \enc, driven by its similarity with previously encountered
objects. Starting from $\obj_e$, it interacts with the user to find
the right position of the encounter, possibly updating the hierarchy
during the process. First, the user could say that $\obj_e$ is not a
\genus of \enc (meaning that their common \genus is further up in the
hierarchy). If this is the case, the algorithm goes up through the
hierarchy until it finds a valid \genus for the encounter. 
This is refined by further interacting with the user via the \textsc{refineGenus} procedure. The two procedures are detailed below.

\begin{algorithm}[t]
	\raggedright
    \begin{algorithmic}[1]
      \Function{PredictGenus}{$\enc$}
          \State $ \obj_e \gets \Call{getRoot}{\calH}$
          \State $ p_e \gets 1.0$
          \For {$v \in \enc$}
                \State $\obj_v, p_v = \Call{PredictVOGenus}{v, \Call{getRoot}{\calH}, 1.0}$
                \If {$\obj_e =  \Call{getRoot}{\calH} \lor p_e < p_v$}
                    \State $ \obj_e \gets \obj_v$
                    \State $ p_e \gets p_v$
                 \EndIf
          \EndFor
          \State  \Return $\langle \obj_e, p_e \rangle$
      \EndFunction
      \State
      \Function{PredictVOGenus}{$v, \obj_v, p_v$}
            \State $\lambda \gets \Call{getRejectionTreshold}{\calK}$
            \State ${\cal C} \gets \Call{getChildren}{\obj_v}$
            \State $p_{c^*} \gets \,max_{\,\obj_c \in {\cal C}} \; \Call{Probability}{\obj_c, v}$
            \State $\obj_{c^*} \gets arg\,max_{\,\obj_c \in {\cal C}} \; \Call{Probability}{\obj_c, v}$
            \If{$p_{c^*} > \lambda$}
                \State  \Return $\Call{PredictVOGenus}{v, \obj_{c^*} , p_{c^*}}$
            \Else
                \State  \Return $\langle \obj_v, p_v \rangle$
            \EndIf
      \EndFunction
    \end{algorithmic}
    \caption{The procedure predicting \genus of an encounter.}
    \label{algo:predictgenus}
\end{algorithm}

\paragraph{Genus prediction.} The \textsc{PredictGenus} procedure outlined in \cref{algo:predictgenus} continuously performs open-world recognition over the evolving \hierarchy of objects seen so far by the machine. The task is to predict the most specific node $\obj_e$ in \hierarchy of the current encounter \enc. The algorithm computes also the probability $p_e$ that \enc belongs to $\obj_e$.
For every visual object $v$ of the current encounter, the candidate \genus is identified. Then, the algorithm outputs the one with maximal probability (excluding the root node, which has always probability 1.0).
The procedure \textsc{PredictVOGenus} in \cref{algo:predictgenus}
computes the \genus of a visual object by navigating down the
hierarchy. By leveraging the notion of \genus and
\differentia, the algorithm searches for the most specific node that
represents the current visual object. Starting from the root of the
hierarchy, it computes the most probable genus child $\obj_{c^*}$ for
the visual object. If its probability $p_{c^*}$ exceeds a rejection
threshold $\lambda$, the procedure recurs over it, otherwise it stops
returning its parent as the \genus of the visual
object. Following~\cite{erculiani2019continual}, the rejection
threshold is chosen with an optimization procedure that maximizes the
number of correct predictions given the previous feedback by the user
(stored in a supervision memory $\calK$). See the original paper for
the details. The procedure terminates when the most probable
node is either below the threshold or is a leaf.

The probability of a \genus being the genus of a visual object can be
formalized as the probability that an element belongs to a set (the
set of previous visual objects for that \genus).  To compute this inclusion probability, we employed the
Extreme Value Machine (EVM) framework~\cite{rudd2018extreme}, because
of its soundness and practical effectiveness. Basically each node is
associated with a set of examples, called \emph{extreme vectors}, that
are used as representatives of the corresponding \genus. We use as
extreme vectors all the visual objects associated with any of the
nodes in the subtree of the node. The probability for a new visual
object is computed based on the closest extreme vector and its
associated probability distribution (a Weibull distribution).

\begin{figure}[!h]
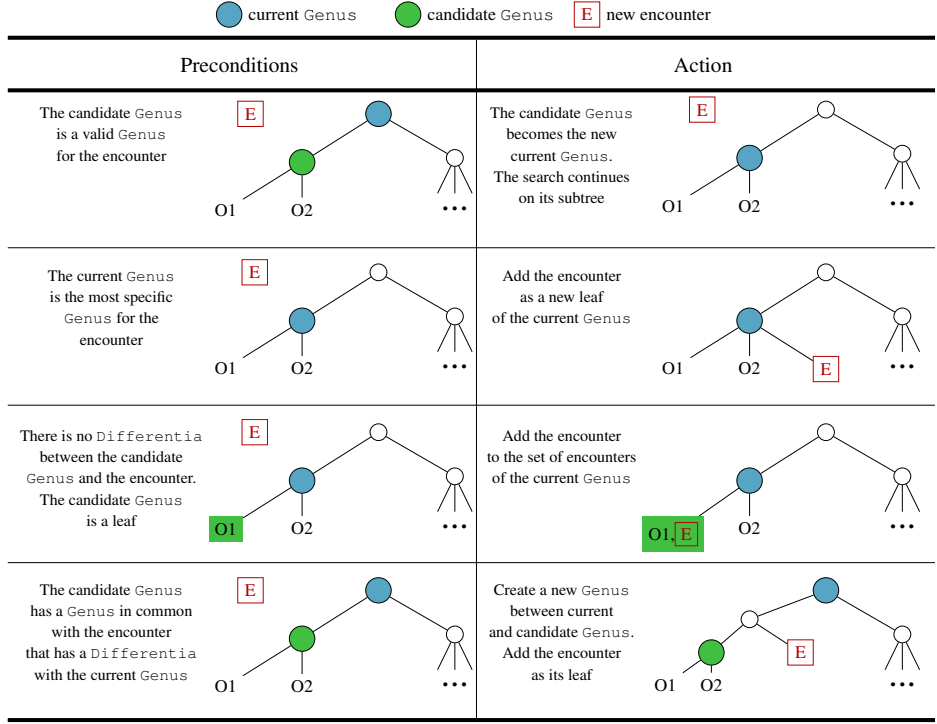

    \centering
    \includestandalone[width=\textwidth]{figures/algocases_andrea}
        \caption{Representation of four possible choices that can be 
        taken during the iterative encounter procedure. 
    }
        \label{fig:algocases}
\end{figure}

\paragraph{Genus refinement.} Starting from the most specific \genus
for the encounter $\enc$ that has been identified by the machine {\em and}
confirmed by the user (called the \emph{current} \genus and corresponding
to "thing" in the simplest case), the algorithm traverses the
hierarchy down asking questions to the user to further refine the
\genus for $\enc$. \cref{fig:algocases} presents a graphical
representation of the four possible situations at each iteration of
the algorithm, and the action to be taken as a result of the user
feedback. 
The new encounter is depicted in red, the current \genus (current best
guess for \genus) in cyan, while the green node is the \genus for
which the machine is asking queries. For each of the possible actions,
each row in the table shows the preconditions that must be met in
terms of \genus and \differentia between the encounter, the candidate
\genus and the current \genus, and the effect that each action has on
the inner hierarchy of the machine. The \textsc{refineGenus} procedure
consists of a sequence of questions and corresponding actions, as
reported in \cref{fig:algocases}, until one of the actions results in
placing the new encounter $\enc$ in the hierarchy (one of the two
lower actions in the figure).

\vspace{-0.2cm}
\section{Experiments}
\label{sec:exp}

\begin{figure*}
	\centering
        \includegraphics[width=0.6\textwidth]{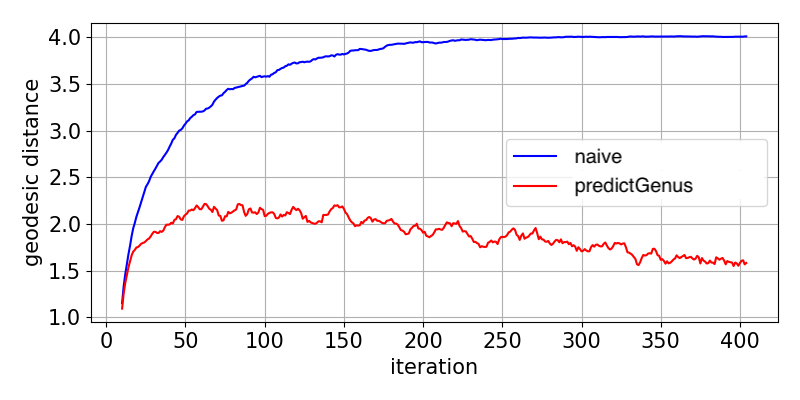}
        \caption{\label{fig:predenc}Comparison between \emph{naive} model and \textsc{PredictGenus} in terms of geodesic distances between the predicted and the correct \genus.}
\end{figure*}

The goal is to evaluate how much the hierarchy built by the machine is aligned with the user hierarchy. This evaluation is done by measuring the distance in the hierarchy between the predicted \genus and the user desired \genus. The greater the distance, the greater the misalignment.
%
All experiments were implemented in
Python3 and PyTorch. 

\paragraph{Data set.} We used a collection of objects organized in 
a perfectly balanced hierarchy of 4 levels, such that each node (except for the leaves)
inside the hierarchy has 3 children, leading to a total of  $3^4 = 81$ leaves.
Each object was recorded 5 different times while rotated or deformed against
a uniform background, thus obtaining $405$ encounters.
The hierarchy was used to simulate the supervision of the user.



\paragraph{Experimental details.} The whole set of videos in the
dataset is presented to the machine in random order. 
An agent simulates the user and
provides supervision to the model by comparing the ground-truth
hierarchy of the data set against the hierarchy of the machine, and
replying to the queries of the algorithm accordingly (see
\cref{fig:algocases}). The hierarchy of objects that is built over
time is always consistent with the ground truth.  The machine goal is
to minimize the categorization effort required from the user when a new
encounter must be placed into the hierarchy. The machine suggests the
starting node, from which the user navigates down the hierarchy until
the correct node is found. The closer the prediction of the machine
is to the ground-truth node, the lower the effort of the user. 

The performance is evaluated in terms of \emph{geodesic distance},
namely the number of edges in the shortest path between the predicted
node and the target node selected by the user. Even if this measure is
affected by the size of the hierarchy (the deeper the tree, the
greater the average distance between couples of nodes), due to the
fact that the evolution of the hierarchy is completely guided by the
user, any model has always its hierarchy updated in the same way. This
fact keeps this performance measure unbiased in the context of this
experiment. We compare the performance  \textsc{PredictGenus}
with that of a \emph{naive} algorithm that
always suggests the root of $\calH$ as the starting node.

\paragraph{Results.}
\cref{fig:predenc} reports the geodesic distance when varying the
number of iterations of the algorithm, averaged
over 100 runs with different random orderings of the objects. Our model, shown in red, substantially
outperforms the naive algorithms (in blue), with the difference becoming more
pronounced as the number of observed objects increases. After an
initial phase in which the cost increases due to the rapid expansion
of the hierarchy, the average geodesic distance for
\textsc{PredictGenus} starts to decrease (at roughly 60
iterations). The algorithm suggests starting nodes closer to the
correct node with higher accuracy because the increasing amount of
encounters allows to better model each \genus. In contrast, the
naive algorithm converges toward an average cost that is equal to the
average distance between the root node and the leaves (the hierarchy
of the dataset is composed by four levels).  Notice that a
particularly bad predictor could in principle do worse than the naive
algorithm, by predicting a node in a subtree that is more distant than
the root node. Albeit preliminary, these results confirm the potential
of the proposed framework in correctly acquiring the hierarchy of the
user and its semantics in terms of \genus and \differentia.

\vspace{-0.2cm}
\section{Related work}
\label{sec:rel}

Our work implements an egocentric, incremental object recognition. A closely related area is continual learning, which addresses the problem of learning to recognize novel objects without forgetting previous knowledge~\cite{de2021continual}.
Traditional machine learning assumes a closed-world setting, where the set of classes is defined at training time. 
Open set approaches reject examples belonging to \emph{unknown unknown} classes~\cite{scheirer2012toward}. Studies on open world recognition extend open set by incrementally updating the model to incorporate the new classes~\cite{bendale2015towards,de2016online,geng2020recent}.
%


In many real-world tasks, such as gene classification and music genre recognition~\cite{simeone2017hierarchical}, the target labels have hierarchical relationships. In computer vision, \cite{lee2018hierarchical,ruiz2022hierarchical} tackle hierarchical novelty detection by identifying to which node in the hierarchy the novel class is attached. Hierarchical information has been used to achieve more reasonable classification errors~\cite{deng2012Hedging,bertinetto2020making} or integrated into neural networks~\cite{deng2014large,nauta2021neural}. These works differ in that do not consider the semantic and egocentric aspects. 


In the works on visual-semantic embeddings, the idea is to map the input feature space to a semantic embedding space~\cite{fu2017vocabulary,zhao2017open}, for instance by projecting the images and the knowledge graph into a unified representation~\cite{lonij2017open}. \cite{farhadi2009describing} learns object attributes, both semantic (part of the objects) and non-semantic (visual feature space), from annotations to classify images. These approaches differ in that they neither try to align recognition with lexical semantics nor use hierarchical classifications.

The approaches that study the grounding of human language in perception, especially vision~\cite{baroni2016grounding}, are strongly related to our work. Examples in this field are answering questions grounded on visual images~\cite{bernardi2021linguistic}, image captioning~\cite{hodosh2013framing}, visual commonsense~\cite{zellers2019recognition} and visual reasoning with natural language~\cite{suhr-etal-2019-corpus}. These approaches do not leverage the work done in lexical semantics to drive the object recognition process.

\vspace{-0.2cm}
\section{Conclusion}
\label{sec:con}

In this paper, we have introduced a novel approach where objects are recognized following the same hierarchical process that is used in lexical semantics to provide meaning to the nouns used to name objects. The first set of experiments shows a consistent improvement in the alignment between what the system recognizes and the words used by humans to describe what is being recognized.

\vspace{-0.2cm}
\section*{Acknowledgments}

This research has received funding from the European Union's Horizon 2020 FET Proactive project ``WeNet - The Internet of us'', grant agreement No.  823783, and from ``DELPhi - DiscovEring Life Patterns'' project funded by the MIUR Progetti di Ricerca di Rilevante Interesse Nazionale (PRIN) 2017 -- DD n. 1062 del 31.05.2019.  The research of AP was partially supported by TAILOR, a project funded by EU Horizon 2020 research and innovation programme under GA No 952215.

\bibliography{reference}
\bibliographystyle{vancouver}

\end{document}

%% file: figures/tikzcommon.tex
\usetikzlibrary{calc}
\usetikzlibrary{math}
\pgfdeclarelayer{background}
\usetikzlibrary{positioning, patterns, fit}
\pgfsetlayers{background,main}
\usetikzlibrary{matrix}
\usetikzlibrary{decorations.pathreplacing,calligraphy}

\tikzset{
    pics/linkedcouple/.style  args={(#1) (#2)}{code={
        \node[pic actions,fit = (#1)] (temp1) {} ;
        \node[pic actions,fit = (#2)] (temp2) {} ;
        \draw (temp1) -- (temp2);
    }},
    matchstyle/.style={draw, green!60!black, line width=.5mm, inner sep =1mm, rounded corners=2mm},
    nomatchstyle/.style={matchstyle, red!40!gray},
    dbgpoint/.style={circle,inner sep=0pt,minimum size=2pt,fill=red},
	keypoint/.style={inner sep=0pt,minimum size=0pt},
    picframe/.style={inner sep=0pt,minimum size=6mm},
	enc/.style={draw,rounded corners=2mm, inner sep= 4mm},
	genus/.style={draw,dashed, line width=0.3mm, rounded corners=2mm, inner sep= 2mm},
	worm/.style={draw,dash dot, rounded corners=2mm, draw opacity=0.5, line width=0.4mm, inner sep =4mm},
    missingimg/.style={fill=gray!10!white, text=black,text width =1cm,  minimum size = 18mm, path picture={
        \draw[gray!30!white]
               (path picture bounding box.south east) -- (path picture bounding box.north west) 
               (path picture bounding box.south west) -- (path picture bounding box.north east);
         \node[anchor=center, gray!60!white, text width=1cm] at (path picture bounding box.center) {missing image} ;
          } 
     }
}

%% file: figures/music/music.tex
\def\disp{0.3}
\begin{tikzpicture}[
    genst/.style={draw,rounded corners=1mm, inner sep= 0mm},
    basenode/.style={draw,text width=3.9cm},
    ]

    \coordinate (root) at (0,0) ;
    
    \node at (root) {\includegraphics[width=15mm]{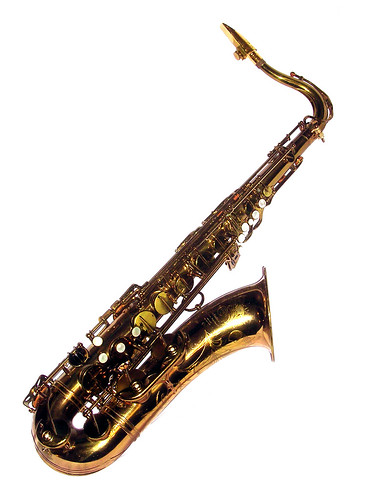}}
    [sibling distance = 3.5cm, level distance = 1.9cm]
        child { node {\includegraphics[width=15mm]{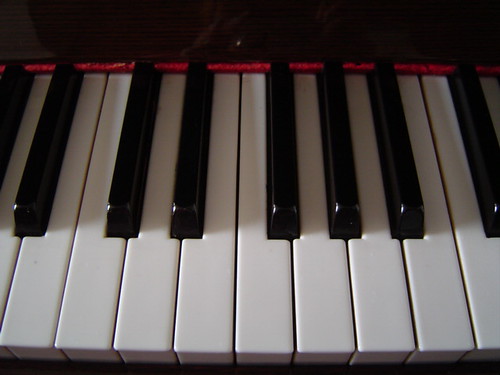}}}
        child { node {\includegraphics[width=15mm]{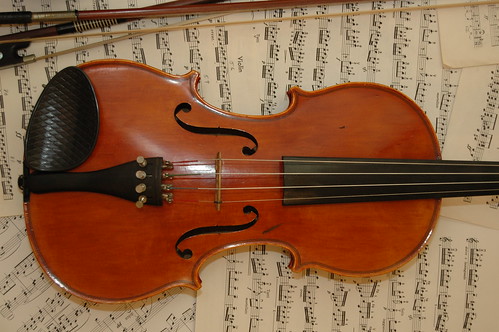}}
                child {node {\includegraphics[width=15mm]{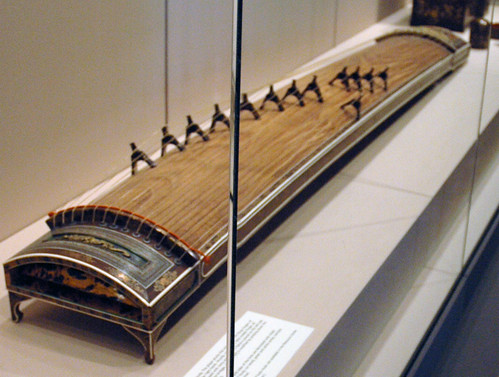}}}
                child {node {\includegraphics[width=15mm]{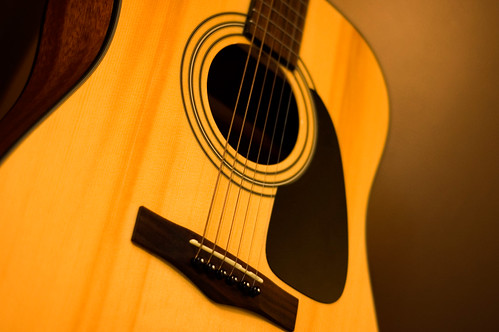}}
                       child {node {\includegraphics[width=15mm]{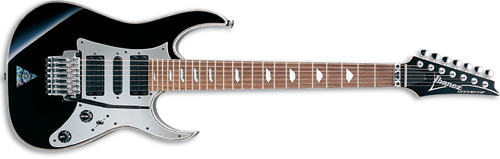}}}
                       child {node {\includegraphics[width=15mm, height=20mm]{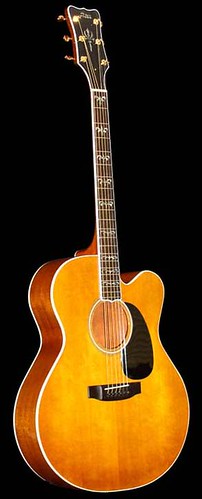}}}
                       }
                child{node {\includegraphics[width=15mm]{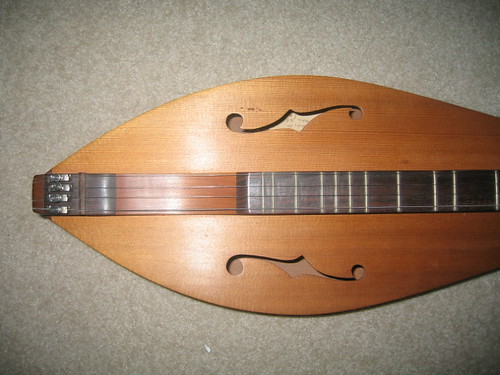}}}
         }
         child {node {\includegraphics[width=15mm]{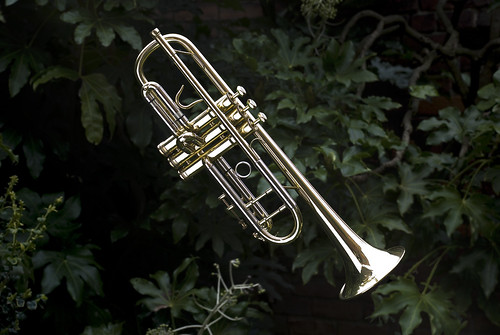}}};
         
    [font=\normalsize]\node[basenode,left= 12 of root] (taxonomy) {\textbf{Musical instruments}\\G: device\\D: with sound mechanism}
    [sibling distance = 5.6cm, level distance = 1.9cm]
        child { node[basenode] {\textbf{Keyboard instrument}\\G: musical instrument\\D: with keyboard}}
        child { node[basenode] {\textbf{String instrument}\\G: musical instrument\\D: with taut strings}
                child {node[basenode] {\textbf{Koto}\\G: string instrument\\D: with 13 strings}}
                child {node[basenode] {\textbf{Guitar}\\G: string instrument\\D: with 6 strings}
                       child {node[basenode] {\textbf{Electric guitar}\\G: guitar\\D: with input jack }}
                       child {node[basenode] {\textbf{Acoustic guitar}\\G: guitar\\D: without input jack}}
                       }
                child{node[basenode] {\textbf{Dulcimer}\\G: string instrument\\D: elliptical body, 3 or 4 strings}}
         }
         child {node[basenode] {\textbf{Wind instrument}\\G: musical instrument\\D: with embouchure}};

\end{tikzpicture}

%% file: figures/hyp2.tex
\begin{tikzpicture}[]

    \tikzmath{
    \lastimg = 7;
    \imsz= 10;
    \reprsz = 20;
    }

    \node[picframe] (0) {\includegraphics[width={\imsz}mm]{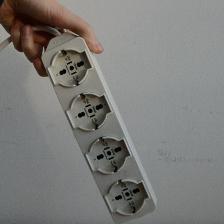}} ;

    \edef\points{(0)}
    \foreach \i in {1,..., \lastimg}{
        \node[picframe, right = 0.1 of {\the\numexpr\i -1\relax}] (\i)  {\includegraphics[width={\imsz}mm]{images/hyp2/\i}};
        \xdef\points{(\i) \points}
    }

        \foreach \i/\o [count=\cnt] in {0/3, 3/6, 6/\lastimg} {
            \node[picframe,  below left= 1.1 and -1.2 of \i, label=below:$v^\cnt$] (o\i) {\includegraphics[width={\reprsz}mm]{images/hyp2/\i}} ;
        \begin{pgfonlayer}{background}
            \draw[->, shorten >= 8, shorten <= 1] ($ (\i.south) - (0.2, 0) $) -| (o\i)  ;
            \node[keypoint, below left = 0.6 and -1.2 of \i] (a\i) {}; 
            \tikzmath{
                \disp = ( 4 - 2*Mod(\cnt,2)) / 15 + 0.4 ;
            }
            \coordinate (a\i) at ($ (\i.south) - (0.3, \disp) $); 
            \draw (\o) |- (a\i) ;
        \end{pgfonlayer}
        }

        \begin{pgfonlayer}{background}
            \node[fill= black!75!white, inner xsep=1mm, inner ysep=3mm, fit= (0) (\lastimg)] (film) {} ;

         \tikzmath{
         \c = (\lastimg + 1) * ( \imsz + 1) + 3;
         int \n;
         \n = ceil(\c / 4) ;
         }

        \foreach \i [evaluate=\i as \si using \i * .4] in {0,..., \n}{ 
        \filldraw[white] ($ (0.north west) + (0,0.1) + (\si, 0) $) 
                        rectangle ($ (0.north west) + (0.2,0.2) + (\si, 0) $) ;
        \filldraw[white] ($ (0.south west) + (0,-0.1) + (\si, 0) $) 
                        rectangle ($ (0.south west) + (0.2,-0.2) + (\si, 0) $) ;
        }
        \end{pgfonlayer}

        \node[worm, yshift=-2mm, inner xsep =2mm,fit=(o0) (o6),label=right:$\enc$] {} ;

\end{tikzpicture}

%% file: figures/hyp1.tex
\begin{tikzpicture}[] 
	\node[picframe] (a) {\includegraphics[width=18mm]{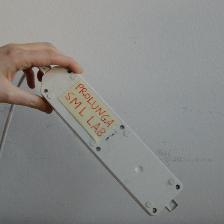}};
	\node[picframe] (b) [right= 0.5 of a] {\includegraphics[width=18mm]{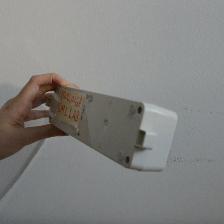}};
	\node[picframe] (c) [right= 0.5 of b] {\includegraphics[width=18mm]{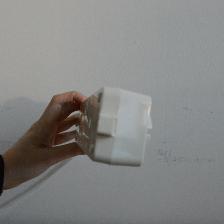}};
    \node[keypoint] (labc) [right= .2 of c] {\Large $\enc_1$};

	\node[picframe] (1) [below=  1 of a]{\includegraphics[width=18mm]{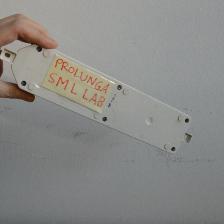}};
	\node[picframe] (2) [right= 0.5 of 1] {\includegraphics[width=18mm]{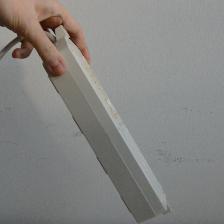}};
	\node[picframe] (3) [right= 0.5 of 2] {\includegraphics[width=18mm]{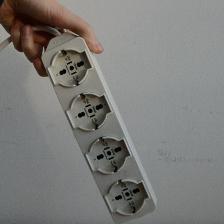}};
    \node[keypoint] (l123) [right= .2 of 3] { \Large $\enc_2$};

    \pic[matchstyle] {linkedcouple={(a) (1)}};

    \node[enc, inner sep =7mm, fit= (l123)(labc)(a) (b) (c) (1) (2) (3),label=left:$\obj_S$] {} ;
    \node[worm, inner sep = 3mm, fit=(labc) (a) (b) (c)] {} ;
    \node[worm, inner sep = 3mm,fit=(l123) (1) (2) (3)] {} ;
\end{tikzpicture}

%% file: figures/hyp3.tex
\begin{tikzpicture}[] 
	\node[picframe] (a) {\includegraphics[width=18mm]{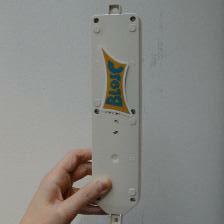}};
	\node[picframe] (b) [right=0.8 of a] {\includegraphics[width=18mm]{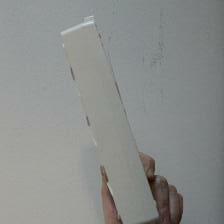}};
	\node[picframe] (c) [right=0.8 of b] {\includegraphics[width=18mm]{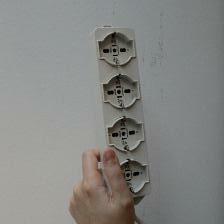}};

	\node[picframe] (1) [below=  1.2 of a]{\includegraphics[width=18mm]{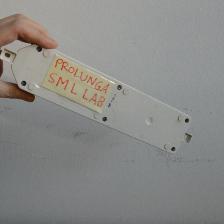}};
	\node[picframe] (2) [right= 0.8 of 1] {\includegraphics[width=18mm]{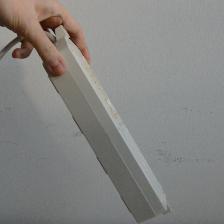}};
	\node[picframe] (3) [right= 0.8 of 2] {\includegraphics[width=18mm]{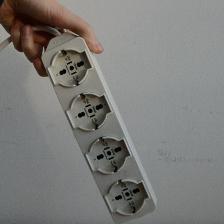}};

    \node[nomatchstyle, fit =(1)] {};
    \node[nomatchstyle, fit =(a)] {};
    \pic[matchstyle] {linkedcouple={(b) (2)}};
    \pic[matchstyle] {linkedcouple={(c) (3)}};

    \node[enc, fit =(a) (b) (c), inner sep = 3mm, label= right:$\obj_2$] {};
    \node[enc, fit =(1) (2) (3), inner sep = 3mm, label= right:$\obj_1$] {};
\end{tikzpicture}

%% file: figures/algocases_andrea.tex
\def\disp{0.3}
\def \SM{$\setminus$}
\begin{tikzpicture}[baseline=(current bounding box.center),
                    parstl/.style={fill=cyan!50!gray, minimum size =5mm},
                    childstl/.style={fill=green!50!gray, minimum size =5mm},
                    nepewstl/.style={fill=yellow, minimum size =5mm},
                    newstl/.style={draw, red!70!black, minimum size =5mm},
                    basenode/.style={draw, circle},
                    newedgestl/.style={red!70!black, dotted}
                    ]   
    \coordinate (root) at (0,0) ;

  \node[circle,draw=black,left= 4.5 of root, parstl,label=right:current \genus] (parent){} ;
    \node[circle,draw=black,right= 3 of parent, childstl, label=right:candidate \genus] (child) {} ;
    \node[right=  3 of child, newstl,
          label=right:new encounter] (new) {E} ;

    \matrix[below = 1.5 of root, row sep =.5cm] (m){
        \node[align=center] (leftmost) {\small The candidate  \genus \\ \small  is a valid \genus \\ \small for the encounter} ; &
        \node[basenode, parstl] (root11) {}[level distance = 0.7cm] 
            child {node[childstl,basenode] {} 
                child {node {O1}}
                child {node {O2}}
                child {node  {} edge from parent[draw=none]}
                }
            child {node {} edge from parent[draw=none]}
            child{
                node[basenode] {}[sibling distance=0.4cm]
                  child {node  {}}
                  child {node (middle) {}}
                  child {node  {}}
             };

         \node at (middle){\huge ...};
         \node[newstl, left=2 of root11] {E}; 
         & 
        \node[align=center] {\small The candidate \genus \\ \small becomes 
                            the new \\ \small current \genus. \\ \small The search continues \\ \small on its subtree} ; 
         &
        \node[basenode] (root12) {}[level distance = 0.7cm] 
            child {node[basenode,parstl] {} 
                child {node {O1}}
                child {node {O2}}
                child {node  {} edge from parent[draw=none]}
                }
            child {node {} edge from parent[draw=none]}
            child{
                node[basenode] {}[sibling distance=0.4cm]
                  child {node  {}}
                  child {node (middle) {}}
                  child {node  {}}
             };

         \node at (middle){\huge ...};
         \node[newstl, left=2 of root12] {E}; 

         \\
        \node[align=center] {\small The current  \genus \\ \small is the most specific \\ \small \genus for the \\ \small encounter}; &
        \node[basenode] (root21) {}[level distance = 0.7cm] 
            child {node[parstl,basenode] {} 
                child {node {O1}}
                child {node {O2}}
                child {node  {} edge from parent[draw=none]}
                }
            child {node {} edge from parent[draw=none]}
            child{
                node[basenode] {}[sibling distance=0.4cm]
                  child {node  {}}
                  child {node (middle) {}}
                  child {node  {}}
             };

         \node at (middle){\huge ...};
         \node[newstl, left=2 of root21] {E}; 
         \node at ($ (root21)  + (0,0.5) $) { } ;
         & 
        \node[align=center] {\small Add the encounter \\ \small  as a new leaf \\ \small  of the current \genus} ; 
         &
        \node[basenode] (root22) {}[level distance = 0.7cm] 
            child {node[basenode,parstl] {} 
                child {node {O1}}
                child {node {O2}}
                child {node[newstl] {E} edge from parent}
                }
            child {node {} edge from parent[draw=none]}
            child{
                node[basenode] {}[sibling distance=0.4cm]
                  child {node  {}}
                  child {node (middle) {}}
                  child {node  {}}
             };

         \node at (middle){\huge ...};
         \\
        \node[align=center] {\small There is no \differentia \\ \small between the candidate \\ \small \genus and the encounter. \\ \small The candidate \genus \\ \small is a leaf} ; &
        \node[basenode]  (root31) {}[level distance = 0.7cm] 
            child {node[parstl,basenode] {} 
              child {node[childstl] {O1}}
                child {node {O2}}
                child {node  {} edge from parent[draw=none]}
                }
            child {node {} edge from parent[draw=none]}
            child{
                node[basenode] {}[sibling distance=0.4cm]
                  child {node  {}}
                  child {node (middle) {}}
                  child {node  {}}
             };

         \node at (middle){\huge ...};
         \node[newstl, left=2 of root31] {E}; 
         \node at ($ (root31)  + (0,0.5) $) { } ;
         & 
        \node[align=center] {\small Add the encounter \\ \small to the set of encounters \\ \small of the current \genus} ; 
         &
        \node[basenode] (root32) {}[level distance = 0.7cm] 
            child {node[basenode,parstl] {} 
                child {node[childstl] {O1,\fcolorbox{red!70!black}{green!50!gray}{\textcolor{red!70!black}{E}}}}
                child {node {O2}}
                child {node  {} edge from parent[draw=none]}
                }
            child {node {} edge from parent[draw=none]}
            child{
                node[basenode] {}[sibling distance=0.4cm]
                  child {node  {}}
                  child {node (middle) {}}
                  child {node  {}}
             };

         \node at (middle){\huge ...};
         \\
        \node[align=center] {\small The candidate \genus \\ \small has a \genus in common \\ \small with the encounter \\ \small that has a \differentia \\ \small with the current \genus} ; &
        \node[basenode,parstl] (root41) {}[level distance = 0.7cm] 
            child {node[childstl,basenode] {} 
                child {node {O1}}
                child {node {O2}}
                child {node  {} edge from parent[draw=none]}
                }
            child {node {} edge from parent[draw=none]}
            child{
                node[basenode] {}[sibling distance=0.4cm]
                  child {node  {}}
                  child {node (middle) {}}
                  child {node  {}}
             };

         \node at (middle){\huge ...};
         \node[newstl, left=2 of root41] {E}; 
         & 
        \node[align=center] {\small Create a new \genus \\ \small between current \\ \small and candidate \genus. \\ \small Add the encounter \\ \small as its leaf } ; 
         &
        \node[basenode,parstl] (root42) {}[level distance = 0.4cm] 
        child {node[basenode] {}
            child {node[basenode,childstl] {} [sibling distance=0.9cm]
                child {node {O1}}
                child {node {O2}}
                child {node  {} edge from parent[draw=none]}
                }
            child[sibling distance=2cm] {node[newstl]{E}}
               }
            child {node {} edge from parent[draw=none]}
            child[level distance = 0.7cm]{
                node[basenode] {}[sibling distance=0.4cm]
                  child {node  {}}
                  child {node (middle) {}}
                  child {node  {}}
             };

         \node at (middle){\huge ...};
         \\
         & & & \\};

         \coordinate (len) at ($ (m.north east)  -  (m.north west) $);
         \coordinate (h) at ($ (m.north east) - (m.south east)  - (0,0.2)$) ;
         \coordinate (rsz) at ($ - 1/4*(h) $) ;
         \draw[line width=0.8mm] (m.north west) -- +(len); 
         \draw (m.north west) ++ (rsz) -- + (len) ++ (rsz) -- + (len)  ++ (rsz) -- + (len)  ++ (rsz) -- + (len) ; 
         \draw[line width=0.8mm] ($ (m.north west) - (h) $) -- +(len); cyan!50!gray

         \node[align=center] (leg) at ($ (m.north west) + 1/4*(len) + (0,0.5) $) {\large Preconditions} ;
         \node[align=center] at ($ (leg) + 1/2*(len) $) {\large Action} ;
         \draw[line width=0.8mm] ($ (m.north west) + (leg.north) + (0,0.5) - (leg.south) $) -- +(len); 

         \draw ($ (m.north west) + 1/2*(len)  + (0.1, 0) $) -- +($ -1*(h)$);
         \draw ($ (m.north west) + 1/2*(len)  + (0.1, 0) + (0,1) $) -- +(0,-1);

         \coordinate (foffset) at (-3, -3) ;
\end{tikzpicture}